\documentclass[conference,a4paper]{IEEEtran}

\usepackage{amsmath,graphicx}
\usepackage{interval}
\usepackage{todonotes}
\usepackage[utf8]{inputenc}
\usepackage{booktabs}
\usepackage{svg}

\usepackage{amsfonts}
\usepackage{pifont}
\usepackage{multirow}
\usepackage{array}
\usepackage{balance}
\usepackage{hyperref}

\usepackage{color}
\definecolor{purple}{rgb}{1,0,1}
\definecolor{gray}{cmyk}{0,0,0,0.1}
\definecolor{yellow}{cmyk}{0,0,0.6,0}
\definecolor{white}{cmyk}{0,0,0,0}
\definecolor{black}{cmyk}{1,1,1,0}
\definecolor{orange}{cmyk}{0,0.5,0.8,0}
\definecolor{green}{cmyk}{0.7,0.1,1,0}
\definecolor{dgreen}{rgb}{0.0,0.6,0.0} 
\definecolor{dred}{rgb}{0.6,0.0,0.0}   

\newcommand{\mcheck}{\textcolor{dgreen}{\ding{51}}}%
\newcommand{\mcross}{\textcolor{dred}{\ding{55}}}%

\hypersetup{
     colorlinks=true,
     linkcolor=blue,
     filecolor=magenta,      
     urlcolor=cyan,
}

\begin{document}

\title{ Hallucinating Dense Optical Flow from Sparse Lidar for Autonomous Vehicles}

\author{\IEEEauthorblockN{Victor Vaquero,
Alberto Sanfeliu and
Francesc Moreno-Noguer}
\IEEEauthorblockA{Institut de Rob\`otica i Inform\`atica Industrial, CSIC-UPC \\
Llorens i Artigas 4-6, 08028 Barcelona, Spain\\
\{vvaquero,sanfeliu,fmoreno\}@iri.upc.edu
}}

\maketitle

\begin{abstract}
In this paper we propose a novel approach to estimate dense optical flow from sparse lidar data acquired on an autonomous vehicle. This is intended to be used as a drop-in replacement of any image-based optical flow system when images are not reliable due to e.g. adverse weather conditions or at night. In order to infer high resolution 2D flows from discrete range data we devise a three-block architecture of multiscale filters  that   combines multiple intermediate objectives, both in the lidar and image domain.  To train this network we  introduce a  dataset with approximately 20K lidar samples of the Kitti dataset which we have augmented with a pseudo ground-truth image-based  optical flow computed using FlowNet2. We demonstrate the effectiveness of our approach on Kitti, and show that despite using the low-resolution and sparse measurements of the lidar, we can regress dense optical flow maps which are at par with those estimated with image-based methods.

\end{abstract}

\IEEEpeerreviewmaketitle

\section{Introduction}


Estimating optical flow (i.e. pixel-level motion) between two consecutive images, has been a long-standing and crucial problem in computer vision~\cite{sun2014quantitative}. It gains special importance for autonomous driving, as it has become an important mid-level feature to then perform higher-level tasks such as  object detection, motion segmentation or time-to-collision estimation.

Yet, it has not been until recently when close to real-time methods providing dense and accurate optical flow from RGB images have appeared~\cite{dosovitskiy2015flownet,ilg2017flownet,sun2017pwc}. These recent approaches are mostly based on deep learning techniques and more specifically convolutional neural networks (CNN), which have shown its capacity to learn more useful features than traditional hand-made variational approaches~\cite{revaud2015epicflow}, providing robustness against rotation, translation and illumination changes.

However, standard RGB cameras may still suffer under harsh weather conditions such as heavy fog, rain or snow. In the autonomous driving field, circumstances can be even harder, as for example during night drives, where other vehicles lights may dazzle our vision systems spoiling the captured images. 
In order to assure full security and protect against sensor temporal failure, self-driven vehicles should be equipped not only with cameras but also with a number of different sensors such as lidar or radar providing diverse and redundant information. 
Sensor fusion, as the ability of merging information from different sources, is therefore a must for these autonomous systems. If we want self-driven vehicles to move and interact with our dynamic and challenging road and urban scenarios, they must contain algorithms that are able to integrate redundant information from different sources.

\begin{figure}[t]
\centering
 \centerline{
  	\includegraphics[width=1\columnwidth]{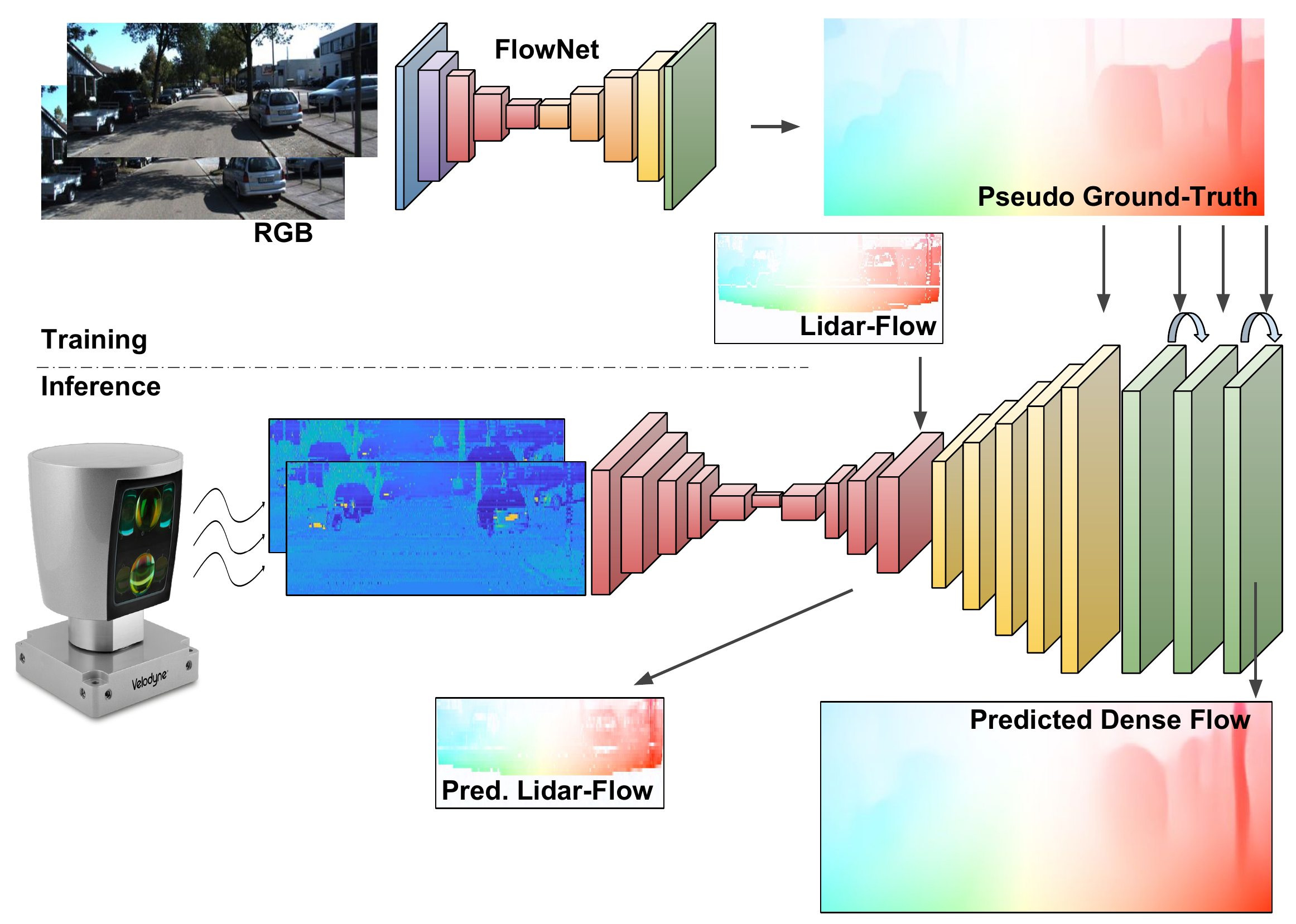}}
  
 \caption{{\bf Dense optical flow from sparse lidar.} We introduce a deep architecture that given two consecutive low-resolution  and sparse lidar scans, produces a high-resolution and dense optical flow, equivalent to one that would be computed from images. Our approach, therefore, can replace RGB cameras  when the quality images is poor due to e.g. adverse weather conditions. Notice that the RGB images shown in the top-left  are only considered to generate the pseudo ground-truth used during training. Inference is done  from only lidar scans. }
  \label{fig:init}
  \vspace{-4mm}
\end{figure}

In this paper we propose a novel deep learning approach based on CNNs which, using only sparse lidar information as input,  is able to estimate in real-time dense and high resolution optical flow that is directly compatible with any camera-based estimated flow. In order to guide the network from the low-resolution and scarce lidar input to the final output we propose an three-block architecture that introduces intermediate learning objectives at different resolutions in both lidar and image domains, as well as refines the obtained prediction increasing the sharpness of the final solution.

One of the main problems that we need to tackle is the lack of training data with corresponding pairs of lidar measurements and image-based optical flow. For training  image-to-optical flow networks, this has commonly been addressed by using synthetically generated datasets~\cite{dosovitskiy2015flownet,Butler2012eccv,MIFDB16flyingThings}. However, virtual datasets that contain optical flow ground-truth do not provide lidar information and thus, are not suitable  for our purposes. On the other hand, real driving datasets (e.g.  the Kitti dataset~\cite{Geiger2012CVPR}) may contain true lidar measurements but not enough corresponding optical flow ground-truth.

To circumvent this lack of training data, we elected a subset of the Kitti dataset (the ``Object Tracking Benchmark'') which is annotated with images and lidar and estimated from the images a high-resolution pseudo ground-truth optical flow using the well established FlowNet2~\cite{ilg2017flownet}. This way, we build a lidar-optical flow dataset with approximately 20K samples.

We provide qualitative evaluation on Kitti and show that, despite feeding our network with low-dimensional and sparse lidar measurements, we are able to predict high-resolution flow maps which are visually appealing (Fig.~\ref{fig:res}). Moreover, we perform also quantitative evaluation on the lidar-available subset of the ``Kitti Flow 2015'' benchmark showing that our approach is on par with other image based regressors and even close to FlowNet2, which is the upper bound we can obtain after generating the used ground-truth optical flow from it.


\section{Related Work}

Optical flow has been used as a source of information for a wide range of  computer vision problems including motion segmentation~\cite{bradski2002motion, vaquero2018deep}, 3D reconstruction~\cite{Trulls_eccv2012}, object tracking~\cite{dang2002fusing} or video encoding~\cite{krishnamurthy1995optical}. 
Initial formulation was proposed by Horn and Schunck in 1981 as a variational approach~\cite{horn1981determining}, aiming to minimize an objective function  with a data term   enforcing  brightness constancy and an spatial term  to model the expected motion fields over the image. Subsequent methods build upon  this scheme  by adding different terms, e.g. combining local and global features~\cite{bruhn2005lucas}, accounting for large displacements~\cite{brox2009large}, or introducing semantic and layered information~\cite{hsu1994accurate, sun2014local}. 
For a more extensive report on classical optical flow methods, the reader is referred to~\cite{sun2014quantitative}.

Although deep learning penetrated with a great force into a number of computer vision problems, its application to build  end-to-end supervised optical-flow systems   was not immediate, basically due to the difficulty of obtaining a sufficiently large training set.
Early convolutional approaches to compute optical flow were focused on improving different parts of the standard pipeline such as the extraction of better non hand-crafted features for patch matching~\cite{revaud2015epicflow, weinzaepfel2013deepflow, guney2016deep}, better segmentations~\cite{bai2016exploiting}, or even layered solutions~\cite{sevilla2016optical}.

The first end-to-end optical flow deep   network was presented in~\cite{dosovitskiy2015flownet}, showing that it was possible to reach state of the art performance training  on synthetic data.
Since then, other CNN architectures have been proposed. In this way, \cite{ranjan2017optical} uses a combination of traditional pyramids and convolutional networks providing features at different resolutions. Contrary, \cite{vvaquero2017flow} introduces an  approach in which fine details are combined to coarse predictions. Recently, \textit{FlowNet2} \cite{ilg2017flownet} has positioned as one of the top performance deep learning optical flow methods. It presents a scheme of stacked CNNs trained separately and with carefully chosen sample-learning schedules for large and small displacements. Following these lines, \cite{sun2017pwc} uses a combination of wrapping techniques and pyramids that, at the time of publication, makes it one of the top performing real-time deep methods in the Kitti flow 2015 benchmark \cite{Menze2015CVPR}.

\begin{figure*}[t!]
\centering
 \centerline{
  	\includegraphics[width=0.95\textwidth]{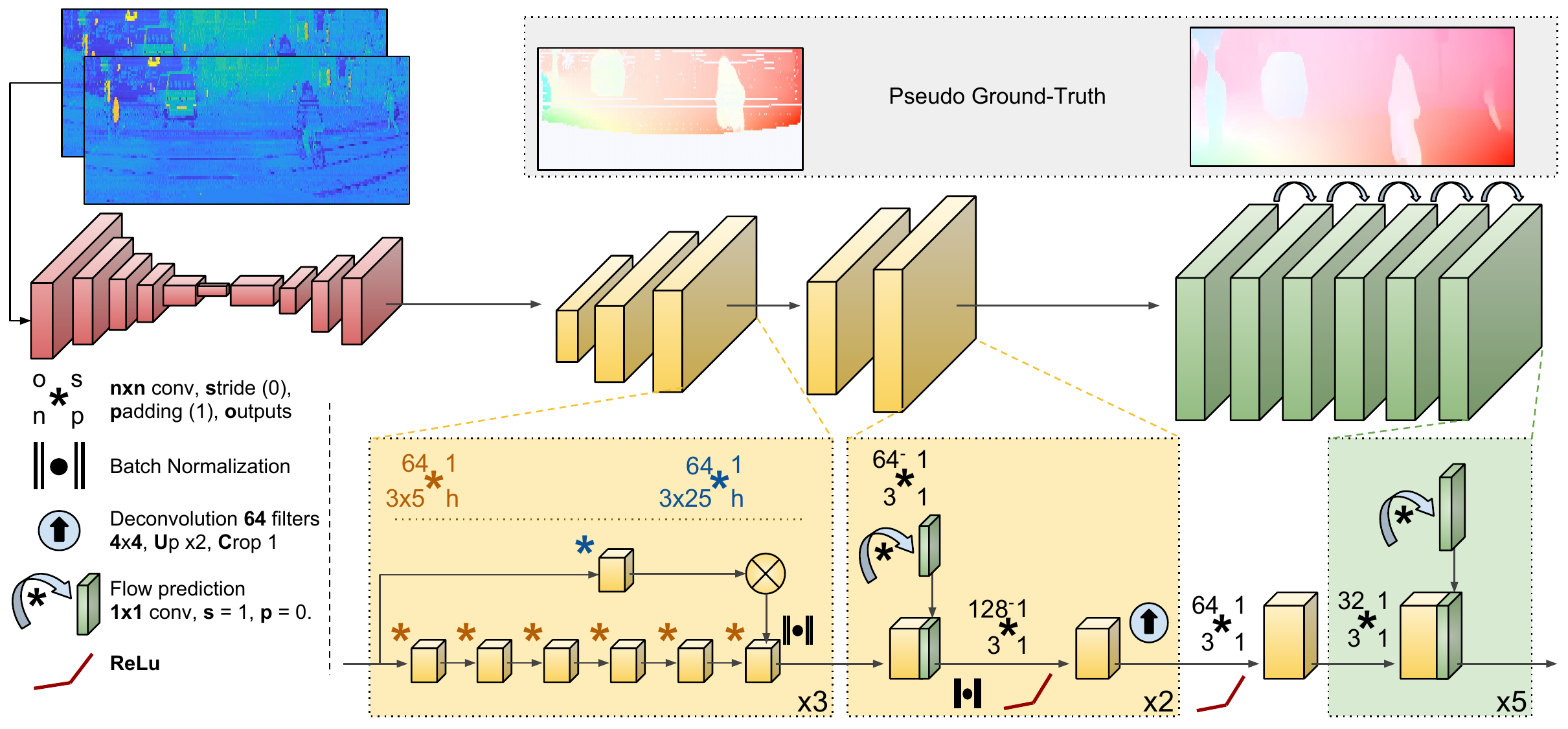}}
  
  \caption{{\bf Lidar to dense optical-flow architecture.} The proposed network is made of three main  blocks sequentially connected which resolve the problem in different stages: 1) Estimation of the lidar-flow in low resolution (red layers); 2) Low-to-high resolution flow transformation and lidar-to-image domain change (yellow layers); 3) Final flow refinement (green layers).} 
 
  \label{fig:approach}
  \vspace{-4mm}
\end{figure*}

Very recent works tackle  the optical flow problem in an unsupervised manner, replacing the supervised loss by a new one that relies on the classical brightness constancy and motion smoothness terms \cite{jason2016back}. Further works on this line make use of more elaborate unsupervised losses taking advantage of warp techniques, as for example in \cite{Meister2018Unflow} where a bidirectional census loss is presented.
These methods alleviate the need of extensive annotated flow ground-truth or the use of virtual environments although they usually need to be very precisely fine-tuned to provide on par results to supervised methods.

Our work also has some connection with the super-resolution literature~\cite{dong2016image}. Nevertheless, note that in super-resolution works, both input and output sources belong to the same type of data. Here, besides having to handle the difference between the input and output spaces and resolution, we need to resolve the additional task of estimating the flow.

\vspace{1mm}
\noindent{\bf Contributions.} All   previous dense optical flow approaches take a pair of  RGB images as input. In this paper we show how a similar high resolution flow can be obtained from a much less informative, but more robust to adverse weather conditions, lidar sensor.  For training our network we  create a lidar-to-image flow dataset, which does neither exist in the literature.


\section{Hallucinating Dense Optical Flow}

We propose a CNN architecture to hallucinate dense high resolution 2D optical flow in the image domain using   as input only sparse and low resolution lidar information. Our network bridges the   gap between lidar and camera domains, so that when the camera images are spoiled (e.g. at night sequences or due to heavy fog), we can still provide an accurate optical flow to directly substitute the degenerated image-based prediction in any vehicle navigation algorithm.

\subsection{General Problem Statement}

Let us define an end-to-end convolutional network to predict dense optical flow as $\mathcal{Y} = \mathcal{F_{\theta}}(\mathcal{X}_t, \mathcal{X}_{t+1}; \theta)$, where $\mathcal{F_{\theta}}$ represents the  network with trainable parameters $\theta$; $\mathcal{X}_t$ and $\mathcal{X}_{t+1}$ $\in \mathbb{R}^{N \times M \times 2}$ are  two consecutive lidar scans (including range and laser reflectivity),  and $\mathcal{Y} \in \mathbb{R}^{H \times W \times 2}$ is the predicted flow represented in a pre-defined image domain of size $H\times W$. Let $[h,w]$ be the position of one pixel within this domain. 

Our problem states two main challenges. On one hand, lidar and image field of view (FOV) are not totally overlapping; On the other hand, the resolution of the input lidar scans $M\times N$ is generally much smaller than the $H\times W$ size of the RGB images on which we seek to hallucinate the flow. A naive end-to-end deep model $\mathcal{F_{\theta}}$ would consist of stacking  convolutions and deconvolutions \cite{zeiler2010deconvolutional} until obtaining the desired $H \times W$ output size in the image FOV.  
However, in the first entry of Table~\ref{tab:results}, we show that a simple model like that it not capable of capturing the correct motion of the scene.

We have therefore devised a more elaborated architecture, as shown in Fig.~\ref{fig:approach}, consisting of three main blocks. 
The first one estimates the motion in the sparse lidar domain using a specific architecture resembling FlowNet~\cite{dosovitskiy2015flownet}, and it is trained with a  ground-truth  lidar optical flow. The second block performs the domain transformation and upsampling, guiding the learning towards predicting the final optical flow in the image domain. Finally, a refinement step is implemented to produce more accurate, dense, and visual appealing predictions. 

In the following sections we first describe the process to create the training data, including the input lidar frames and their associated image-based optical flows.  We then describe each one of the building blocks of the proposed architecture.

\subsection{Input/Output Data}
\label{subsec:data}

Existing CNNs for inferring optical flow (e.g. FlowNet), resort to  synthetic training datasets with complete and dense ground-truth flow~\cite{dosovitskiy2015flownet, Butler2012eccv, MIFDB16flyingThings}. While the rendered images look very realistic, these datasets are not annotated with range nor reflectivity information provided by real laser sensors.

To learn the parameters of  the proposed deep architecture we need the following training data: i) lidar data aligned with an RGB camera, i.e. we need to know the mapping from the 3D range measurements to the image plane on which we aim to densely hallucinate the optical flow; ii) corresponding optical flow ground-truth annotated in the image domain. Both these types of data are by themselves scarce, and there exist no dataset containing both of them put in correspondence. 

\begin{figure}[t!]
\centering
 \centerline{
  	\includegraphics[width=0.95\columnwidth]{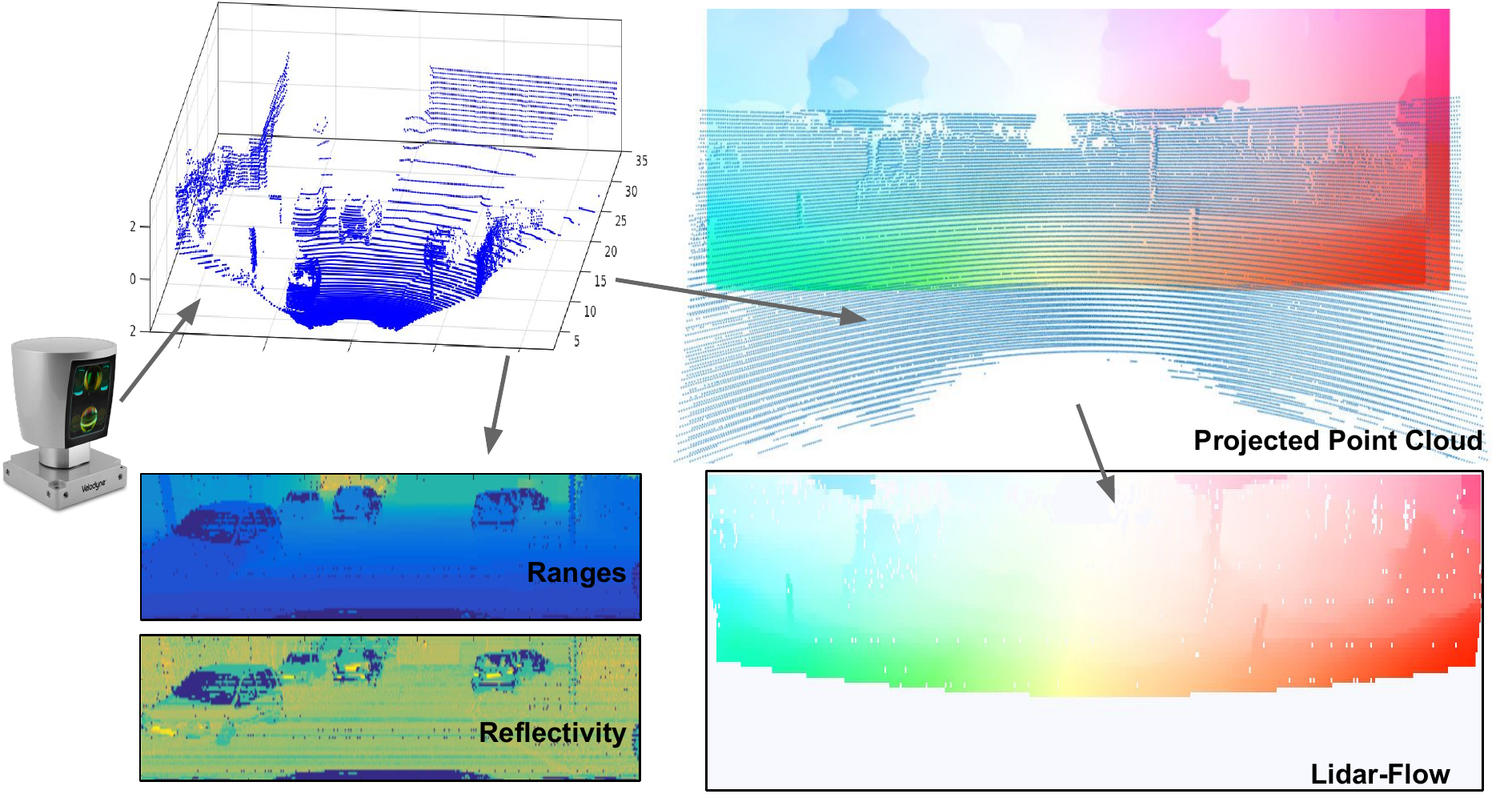}}
  
  \caption{{\bf Building a lidar-to-optical flow dataset.} Given a 3D point cloud from a laser scan (top-left) we create our input tensots with the range and reflectivity information (bottom-left) to be used as our inputs. Lidar-flow pseudo ground-truth is alsoe created by cropping the overlaping areas between the dense image-flow and the projected point cloud (bottom-right).} 
  \label{fig:dataAdapt}
  \vspace{-4mm}
\end{figure}

In order to build the input lidar data we consider the  Kitti Tracking dataset \cite{Geiger2012CVPR}, which specifically provides measurements from a Velodyne HDL-64 sensor, with $64$ lasers vertically arranged rotating at a speed of $10Hz$. 
We first crop the full laser point cloud to obtain the its horizontal overlapping FOV over the camera image. 
Then, we build our $\mathcal{X}_i \in \mathbb{R}^{N \times M \times 2}$ input tensors by projecting the remaining laser points into an $N \times M$ matrix, where for each $[n,m]$ pair $\in [1,N] \times [1,M]$, we encode range and reflectivity values of each corresponding laser beam, as shown in Fig.~\ref{fig:dataAdapt}. More details about this process can be found in \cite{vaquero2017deconvolutional}.

The Kitti Traking dataset we used to extract the lidar measurements is not annotated with optical flow ground-truth. However, there exist an associated RGB image per each lidar scan from which we computed a pseudo ground-truth for optical flow using FlowNet2~\cite{ilg2017flownet}.
Due to the specific  Kitti vehicle's setup, the vertical field of view of the Velodyne sensor does not cover the full corresponding RGB image height $H$. In order to adapt the image-based FlowNet2 predictions to cover the Velodyne vertical FOV we again perform a cropping operation, although this time over the image domain to eliminate those non overlapping areas (see Fig.~\ref{fig:dataAdapt}). Let us denote by $GT_{Dense} \in  \mathbb{R}^{H \times W \times 2}$ this pseudo optical flow ground-truth that we will use for training.

At this point, we have already built a training set consisting of input  lidar frames $\{\mathcal{X}_t,\mathcal{X}_{t+1}\}$ and its corresponding image-based optical flow ground-truth $GT_{Dense}$, which would be enough to train a naive end-to-end regressor as shown in the first entry of Table \ref{tab:results}. 
However, as we will see next, we obtain far better results by building intermediate objectives using subsets of training data as well as domain specific losses.

\subsection{Lidar Flow}
\label{subsec:lidarFlow}
The first block of our proposed architecture aims at predicting \textit{lidar flow}, this is, the low resolution flow in the lidar domain from two consecutive lidar point clouds $\mathcal{X}_t$ and $\mathcal{X}_{t+1}$. The ground truth lidar flow for this problem, $GT_{Lidar}\in  \mathbb{R}^{N \times M \times 2}$, is computed by projecting the lidar point cloud $\mathcal{X}_t$ onto the dense image flow $GT_{Dense}$ and keeping the motion and reflectance values of the overlapping points (See Fig.~\ref{fig:dataAdapt}). Since the input lidar frames are low resolution, noisy and scattered, so will it be $GT_{Lidar}$.

In order to learn this low dimensional flow, we train a network $\mathcal{Y}_{Lidar} = \mathcal{G}_{\theta_{Lidar}}(\mathcal{X}_t, \mathcal{X}_{t+1}; \theta_{Lidar}; GT_{Lidar})$, being $\mathcal{Y}_{Lidar}$ the predicted lidar-flow and $\theta_{Lidar}$ the trainable parameters of the network $\mathcal{G}_{\theta_{Lidar}}$. This network follows a similar contractive-expansive architecture as FlowNet~\cite{dosovitskiy2015flownet}. The main difference w.r.t. FlowNet is that we concatenate up to $5$ contraction levels, creating feature maps of up to $1/32$ of the initial lidar-input resolution.  Moreover, we expanse the feature maps up to half of the initial resolution: $\mathcal{Y}_{Lidar} \in \mathbb{R}^{N/2 \times M/2 \times 2}$, leaving room for the next blocks to perform the hallucination and refinement steps. The shortcuts, intermediate predictions, filter sizes, steps and padding hyperparameters are the same as those used in FlowNet.

\subsection{Lidar to Image Domain Transformation}
\label{subsec:up}

The second major block of the proposed architecture is in charge of bringing the low resolution lidar flow to the high resolution image domain. Specifically, this block  receives as input the lidar flow $\mathcal{Y}_{Lidar}$ predictions along with the accordingly downscaled input lidar frames and produces as output an upscaled image-centered optical flow prediction learned from $GT_{Dense}$. This upscaling operation  can be formally written as  $\mathcal{Y}_{Up} = \mathcal{H}_{\theta_{Up}}(\mathcal{X}^{\prime}_{t}, \mathcal{X}^{\prime}_{t+1}, \mathcal{Y}_{Lidar}; \theta_{Up}; GT_{Dense})$, where $\mathcal{H}_{\theta_{Up}}$ represents the model with learned parameters $\theta_{Up}$, $\mathcal{X}^{\prime}$ refers to the $1/2$ downsampled input lidar frames that match the $\mathcal{Y}_{Lidar}$ resolution, and $\mathcal{Y}_{Up} \in \mathbb{R}^{H \times W \times 2}$ is the output predicted optical flow in the image domain as seen in Fig.~\ref{fig:res}.

In order to actively guide this domain transformation process, we devise an architecture with two sub-blocks (shown in yellow in Fig.~\ref{fig:approach}). 
The first sub-block consists of  a set of multi-scale filters in two convolutional branches, providing context knowledge to the network. In one branch we produce high frequency features by applying $5$ consecutive convolutional layers with small $3 \times 5$ filters and without any lateral padding (which allows the feature maps to grow horizontally for matching the desired output resolution). In the other branch, lower frequency features are generated with a convolution layer using wider $3 \times 25$ filters and outputting the same feature map resolution. Finally, the features of the two branches are concatenated. As can be seen in Fig.~\ref{fig:approach}, this two-branched expansion process is replicated three times. The spatial final resolution of this sub-block feature map is $H/8 \times W/8$ and no flow prediction is performed at this point.
The second sub-block raises the resolution of the feature map to the final size $H/2 \times W/2$ in the image domain, performing iteratively a refinement of the flow. 
This full process is repeated twice until the desired final flow resolution is obtained. Notice here that we upsampled until $H/2 \times W/2$ to speed the system, as in our experiments, including a third block do not produce better results than just bilinearly interpolate the final prediction.

\subsection{Hallucinated Optical Flow Refinement}
\label{subsec:refin}

The flow predicted in the previous step tends to be over-smoothed. Algorithms predicting dense images in which the contours are important (e.g. semantic segmentation) commonly perform refinement steps to produce more accurate outputs. Conditional Random Fields (CRF) is one of the preferred methods for this purpose, and has recently been approached by using Recurrent Neural Networks \cite{dosovitskiy2015flownet, zheng2015conditional, wu2017squeezeseg}. The procedure can be roughly seen as an iterative process over a previous solution. 
We design a similar iterative convolutional approach for refining the hallucinated optical flow prediction, as is sketched in Fig.~\ref{fig:approach}, so that avoiding the computational burden of a CRF and obtaining a fully end-to-end procedure.

We formally denote this final refinement step as $\mathcal{Y}_{End} = \mathcal{K}_{\theta_{End}}(\mathcal{Y}_{Up}; \theta_{End}; GT_{Dense})$. It works by performing a prediction of the final optical flow, which is concatenated to the feature maps of the previous convolutional layer. 
The concatenated tensor is then passed to another block that generates again new feature maps and a new optical flow prediction, but this time with a better knowledge of the desired output. As shown in the green block of Fig~\ref{fig:approach}, this process is repeated 5 times, simulating an iterative scheme.


\section{Experiments}
\label{sec:exps}

\vspace{1mm}
\noindent{\bf Train, test and validation sets.}
As mentioned in Section \ref{subsec:data}, we prioritize the use of real lidar information over having synthetic but accurate optical flow. For this purpose, in our experiments  we use  the KITTI Tracking benchmark \cite{Geiger2012CVPR}, which contains both RGB and lidar measurements from a Velodyne HDL-64 sensor grouped in 50 different sequences, providing us with 19,045 sample pairs for input. 

As pseudo ground-truth for training our architectures, we use the optical flow $GT_{Dense}$ predicted by Flownet2 \cite{ilg2017flownet} from the RGB images,  and the associated   lidar measurements $GT_{Lidar}$, processed as described  in  Section \ref{subsec:lidarFlow}. 
We would like to point that our approach do not consider any RGB image at all during inference, and these are only used to create the pseudo ground-truth optical flow for training.

To provide a  quantitative evaluation for our models, we need real ground-truth to compare against. For this, we use the training set of the KITTI flow 2015 benchmark, which is composed of 200 pairs of RGB images. However, no Velodyne information is provided on these samples, and we therefore had to perform a match search between the RGB images of both KITTI flow and tracking benchmarks. By doing this, we were able to annotate $90$ pairs of Velodyne scans with their associated real optical flow  in the image domain, which compose our \textit{test} set $GT_{Test}$. We split the remaining 18,955 Velodyne frame pairs into two subsets, creating a \textit{train} set of 17,500 samples and a \textit{validation} set of 1455 samples, each of those containing non-overlapping sequences.

\vspace{1mm}
\noindent{\bf Lidar and flow resolutions.} In our experiments, input lidar frames $\mathcal{X}_i$ have a size of $N = 64; M = 384$, where each  entry accounts for the range and reflectivity values measured by the Velodyne sensor. The final output  $GT_{Dense}$ has a resolution of $H = 256; W = 1224$. This increase of the resolution poses the  main difficulty of our approach, which needs to predict dense optical flow with almost thirteen times less input data from a noisy and scarce source.

\vspace{1mm}
\noindent{\bf Implementation details.} Our modules are trained in an  end-to-end manner, in the way that the output of each one becomes the input to the next. For that we use the MatConvNet \cite{vedaldi15matconvnet} Deep Learning Framework. All the weights are initialized with the He's method~\cite{he2015delving} and Adam optimization is used with standard parameters $\beta_1 = 0.9$ and $\beta_2 = 0.999$. 
Training is carried out on a single NVIDIA 1080Ti GPU throughout 400,000 iterations, each iteration containing a batch of 10 pairs of consecutive Velodyne scans. Data augmentation is performed over the lidar inputs only by flipping  the frames horizontally with a 50\% chance, so that preserving the strong geometric properties of the laser measurements and the natural movement of the scene. 
Learning rate is set to $10^{-3}$ for the first 150,000 iterations, and halved each 60,000 iterations. 

\begin{table}[t!]
\centering
\begin{tabular}{llll||c|c|c|c|}
\multicolumn{3}{c}{Modules}                                                                                              &     & \multirow{2}{*}{Fl-BG} & \multirow{2}{*}{Fl-FG} & \multirow{2}{*}{Fl-ALL} & \multirow{2}{*}{EPE} \\ 
\multicolumn{1}{l|}{LF}                 & \multicolumn{1}{l|}{BS}                 & \multicolumn{1}{l|}{RF}                 &     &                                                                              &                                                                              &                                                                               &                      \\ \hline \hline
\multicolumn{1}{l|}{\multirow{2}{*}{\mcross}} & \multicolumn{1}{l|}{\multirow{2}{*}{\mcross}} & \multicolumn{1}{l|}{\multirow{2}{*}{\mcross}} & Noc & 	56.74 	& 	82.75	& 61.24	& \multirow{2}{*}{14.19}                     \\
\multicolumn{1}{l|}{}                  & \multicolumn{1}{l|}{}                  & \multicolumn{1}{l|}{}                  & Occ &	58.11 & 83.14	& 62.04	&                      \\ \hline

\multicolumn{1}{l|}{\multirow{2}{*}{\mcheck}} & \multicolumn{1}{l|}{\multirow{2}{*}{\mcross}} & \multicolumn{1}{l|}{\multirow{2}{*}{\mcross}} & Noc & 23.20	& 	57.67	& 29.15 & \multirow{2}{*}{6.78}                     \\
\multicolumn{1}{l|}{}                  & \multicolumn{1}{l|}{}                  & \multicolumn{1}{l|}{}                  & Occ    & 	24.80 & 58.56	& 30.09 &                      \\ \hline

\multicolumn{1}{l|}{\multirow{2}{*}{\mcheck}} & \multicolumn{1}{l|}{\multirow{2}{*}{\mcross}} & \multicolumn{1}{l|}{\multirow{2}{*}{\mcheck}} & Noc & 20.09    &  54.02 	& 25.95 &  \multirow{2}{*}{5.49}                     \\
\multicolumn{1}{l|}{}                  & \multicolumn{1}{l|}{}                  & \multicolumn{1}{l|}{}                  & Occ    &  22.26  & 54.51 	& 27.31 &					   \\ \hline

\multicolumn{1}{l|}{\multirow{2}{*}{\mcheck}} & \multicolumn{1}{l|}{\multirow{2}{*}{\mcheck}} & \multicolumn{1}{l|}{\multirow{2}{*}{\mcheck}} & Noc & 	18.65 	&  51.56 	& 24.33 &  \multirow{2}{*}{5.16}                    \\
\multicolumn{1}{l|}{}                  & \multicolumn{1}{l|}{}                  & \multicolumn{1}{l|}{}                  & Occ   &  20.88  & 52.58 	& 25.84 &					   \\ \hline        

\multicolumn{3}{c|}{\multirow{2}{*}{FlowNet2\cite{ilg2017flownet} (*)}}  & Noc  & 7.24 & 5.6 & 6.94 & \multirow{2}{*}{-}                      \\
\multicolumn{3}{l|}{}	& Occ  & 10.75 & 8.75 & 10.41 & \\ \hline

\multicolumn{3}{c|}{\multirow{2}{*}{InterpoNet\cite{zweig17interponet} (*)}}  & Noc  & 11.67 & 22.09 & 13.56 & \multirow{2}{*}{-}                      \\
\multicolumn{3}{l|}{}                                                                                                     & Occ                          & 22.15 & 26.03 & 22.80 & \\ \hline

\multicolumn{3}{c|}{\multirow{2}{*}{EpicFlow\cite{revaud2015epicflow} (*)}}  & Noc  & 15.00 & 24.34 & 16.69 & \multirow{2}{*}{-}                      \\
\multicolumn{3}{l|}{}                                                                                                     & Occ                          & 25.81 & 28.69 & 26.29 & \\ \hline

\end{tabular}
\caption{{\bf Quantitative evaluation and comparison with image-based optical-flow methods.} Flow for \textit{background}, \textit{foreground} and \textit{all} is measured for both non-occluded and full points, as in the Kitti Flow benchmark. The End-Point-Error (EPE) is measured against the pseudo ground-truth computed using FlowNet2. (*) indicates that for these methods the test set is slightly bigger from the one used for our approach, as we could no obtain all the corresponding lidar frames. Although indicative, these results show that our lidar-based approach is on par with other well-known optical flow algorithms that rely on higher resolution and quality input images.}
\label{tab:results}
\vspace{-4mm}
\end{table}

\begin{figure*}[t!]
\centering
 \centerline{
  	\includegraphics[width=0.98 \textwidth]{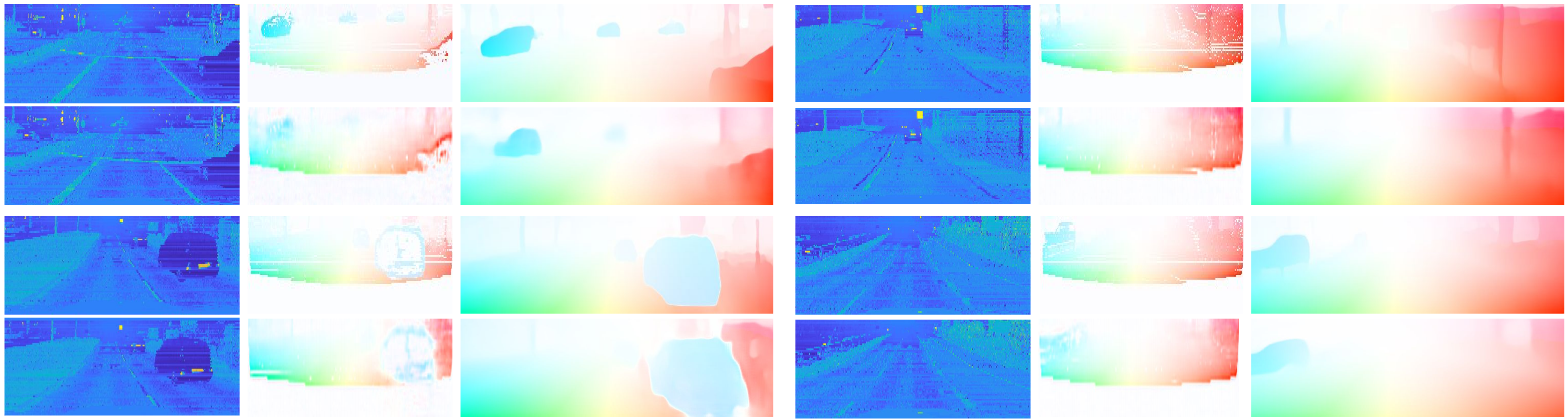}}
  
  \caption{{\bf Qualitative results of our system.} All images are taken during inference from our validation lidar-image-flow set, so none of them were previously seen during training. We show four different example scenes, grouped in 4 quadrants. In each quadrant there are three columns, representing the following. Column 1:  lidar inputs $\mathcal{X}_t$ and $\mathcal{X}_{t+1}$; Column 2-top: lidar flow pseudo ground-truth $GT_{Lidar}$; Column 2-bottom: lidar flow prediction $\mathcal{Y}_{Lidar}$; Column 3-top: dense pseudo ground-truth, $GT_{Dense}$; Column 3-bottom: final predicted dense optical flow $\mathcal{Y}_{End}$. Full sequences: \href{url}{https://youtu.be/94vQUwCZLxQ}}
 \vspace{-3mm}
  \label{fig:res}
\end{figure*}

\vspace{1mm}
\noindent{\bf Definition of the loss.}
Our  approach performs  an end-to-end regression, for which the learning loss is measured at up to twelve places $\sum^{12}_{i=1}{\lambda\mathcal{L}_i}$: five of them in the \textit{Lidar-flow} module described in Section \ref{subsec:lidarFlow}, two more in the upsampling step shown in Section \ref{subsec:up} and the last five at full resolution in the final refinement step detailed in Section \ref{subsec:refin}. 
All these losses compute the L2-norm of the difference between the predicted optical flow and the corresponding pseudo ground-truth of the Training set. The $\lambda$ parameters are set to 1.

\vspace{1mm}
\noindent{\bf Ablation study.} We performed the ablation study summarized in Table \ref{tab:results} to analyse the contributions of the different blocks of our architecture. As quantitative measurements, we follow the Kitti Flow 2015 benchmark guidelines obtaining the Percentage of outlier pixels. A pixel is considered to be correctly estimated if the End-Point-Error (EPE) calculated as the averaged Euclidean distance between the prediction and the real ground-truth $G_{Test}$ is $<3$px or $<5$\%. These measurements are averaged over background regions only, over foreground regions only, and over all ground truth pixels, which respectively are denoted in Table \ref{tab:results} as ``Fl-BG'', ``Fl-FG'' and ``All''. The ``Noc'' and ``Occ'' values refer respectively to the evaluation performed over the Non-Occluded regions and over all the regions. 
In addition, we include the EPE obtained against our Validation set, which give us an idea of how close we are to our upper bound of FlowNet2. 

At the light of the results shown in Table \ref{tab:results}, it is clear that our approach benefits from the inclusion of the different modules. In addition, when comparing with other methods, we can conclude that our full approach performs very close to other state of the art methods which use high-resolution images as input. Although we suffer from larger  errors for the foreground predictions, our overall results are on par with the ones obtained by e.g. EpicFlow~\cite{revaud2015epicflow}, one of the first methods exploiting deep architectures.  Note also the robustness of our approach to occlusions, as the difference between results for both ``Noc'' and ``Occ'' is less significant than in other methods, and even better or very similar to InterpoNet \cite{zweig17interponet} and EpicFlow \cite{revaud2015epicflow}. Some qualitative results are shown on Fig~\ref{fig:res}, including intermediate \textit{Lidar-flow} predictions.

\section{Conclusion}
In this paper we have presented an approach to regress high resolution image-like optical flow from low resolution and sparse lidar measurements. For this purpose, we have designed a deep network architecture made of several blocks that incrementally solve the problem, first estimating a low resolution lidar flow, and then increasing the resolution of the flow to that of the image domain. For training our network we have created a new dataset of corresponding lidar scans and high-resolution image flow predictions, that we use as pseudo ground-truth for training. The results show that the flows estimated by our architecture are competitive with those computed by methods that rely on high-resolution input images. There is still room for improvement in order to get more accurate flow predictions that we left for future work. For example, the addition of better refinement steps to improve results on foreground objects.

\noindent\textbf{Acknowledgment.} 
This work has been supported by the Spanish Ministry of Economy, Industry and Competitiveness projects COLROBTRANSP (DPI2016-78957-R), HuMoUR (TIN2017-90086-R), 
the Spanish State Research Agency through the Maria de Maeztu Seal of Excellence (MDM-2016-0656), 
and the EU project LOGIMATIC (H2020-Galileo-2015-1-687534). 
We also thank Nvidia for hardware donation under the GPU Grant Program.


\bibliographystyle{./IEEEtran}
\bibliography{./IEEEabrv,./references}

\begin{thebibliography}{10}
\providecommand{\url}[1]{#1}
\csname url@samestyle\endcsname
\providecommand{\newblock}{\relax}
\providecommand{\bibinfo}[2]{#2}
\providecommand{\BIBentrySTDinterwordspacing}{\spaceskip=0pt\relax}
\providecommand{\BIBentryALTinterwordstretchfactor}{4}
\providecommand{\BIBentryALTinterwordspacing}{\spaceskip=\fontdimen2\font plus
\BIBentryALTinterwordstretchfactor\fontdimen3\font minus
  \fontdimen4\font\relax}
\providecommand{\BIBforeignlanguage}[2]{{%
\expandafter\ifx\csname l@#1\endcsname\relax
\typeout{** WARNING: IEEEtran.bst: No hyphenation pattern has been}%
\typeout{** loaded for the language `#1'. Using the pattern for}%
\typeout{** the default language instead.}%
\else
\language=\csname l@#1\endcsname
\fi
#2}}
\providecommand{\BIBdecl}{\relax}
\BIBdecl

\bibitem{sun2014quantitative}
D.~Sun, S.~Roth, and M.~Black, ``A quantitative analysis of current practices
  in optical flow estimation and the principles behind them,''
  \emph{International Journal of Computer Vision}, vol. 106, pp. 115--137,
  2014.

\bibitem{dosovitskiy2015flownet}
A.~Dosovitskiy, P.~Fischer, E.~Ilg, P.~Hausser, C.~Hazirbas, V.~Golkov,
  P.~van~der Smagt, D.~Cremers, and T.~Brox, ``Flownet: Learning optical flow
  with convolutional networks,'' in \emph{Proc. ICCV}, 2015.

\bibitem{ilg2017flownet}
E.~Ilg, N.~Mayer, T.~Saikia, M.~Keuper, A.~Dosovitskiy, and T.~Brox, ``Flownet
  2.0: Evolution of optical flow estimation with deep networks,'' in
  \emph{Proc. CVPR}, 2017.

\bibitem{sun2017pwc}
D.~Sun, X.~Yang, M.-Y. Liu, and J.~Kautz, ``Pwc-net: Cnns for optical flow
  using pyramid, warping, and cost volume,'' \emph{arXiv}, 2017.

\bibitem{revaud2015epicflow}
J.~Revaud, P.~Weinzaepfel, Z.~Harchaoui, and C.~Schmid, ``Epicflow:
  Edge-preserving interpolation of correspondences for optical flow,'' in
  \emph{Proc. CVPR}, 2015.

\bibitem{Butler2012eccv}
D.~J. Butler, J.~Wulff, G.~B. Stanley, and M.~J. Black, ``A naturalistic open
  source movie for optical flow evaluation,'' in \emph{Proc. ECCV}, 2012.

\bibitem{MIFDB16flyingThings}
N.Mayer, E.Ilg, P.H{\"a}usser, P.Fischer, D.Cremers, A.Dosovitskiy, and T.Brox,
  ``A large dataset to train convolutional networks for disparity, optical
  flow, and scene flow estimation,'' in \emph{Proc. CVPR}, 2016.

\bibitem{Geiger2012CVPR}
A.~Geiger, P.~Lenz, and R.~Urtasun, ``Are we ready for autonomous driving? the
  kitti vision benchmark suite,'' in \emph{Proc. CVPR}, 2012.

\bibitem{bradski2002motion}
G.~R. Bradski and J.~W. Davis, ``Motion segmentation and pose recognition with
  motion history gradients,'' \emph{Machine Vision and Applications}, vol.~13,
  no.~3, pp. 174--184, 2002.

\bibitem{vaquero2018deep}
V.~Vaquero, A.~Sanfeliu, and F.~Moreno-Noguer, ``Deep lidar cnn to understand
  the dynamics of moving vehicles,'' in \emph{Proc. ICRA}, 2018.

\bibitem{Trulls_eccv2012}
E.~Trulls, A.~Sanfeliu, and F.~Moreno-Noguer, ``Spatiotemporal descriptor for
  wide-baseline stereo reconstruction of non-rigid and ambiguous scenes,'' in
  \emph{Proc. ECCV}, 2012.

\bibitem{dang2002fusing}
T.~Dang, C.~Hoffmann, and C.~Stiller, ``Fusing optical flow and stereo
  disparity for object tracking,'' in \emph{Proc. ITS}, 2002.

\bibitem{krishnamurthy1995optical}
R.~Krishnamurthy, P.~Moulin, and J.~Woods, ``Optical flow techniques applied to
  video coding,'' in \emph{Proc. ICIP}, 1995.

\bibitem{horn1981determining}
B.~K. Horn and B.~G. Schunck, ``Determining optical flow,'' \emph{Artificial
  intelligence}, vol.~17, no. 1-3, pp. 185--203, 1981.

\bibitem{bruhn2005lucas}
A.~Bruhn, J.~Weickert, and C.~Schn{\"o}rr, ``Lucas/kanade meets horn/schunck:
  Combining local and global optic flow methods,'' \emph{International Journal
  of Computer Vision}, vol.~61, no.~3, pp. 211--231, 2005.

\bibitem{brox2009large}
T.~Brox, C.~Bregler, and J.~Malik, ``Large displacement optical flow,'' in
  \emph{Proc. CVPR}, 2009.

\bibitem{hsu1994accurate}
S.~Hsu, P.~Anandan, and S.~Peleg, ``Accurate computation of optical flow by
  using layered motion representations,'' in \emph{Proc. ICPR}, 1994.

\bibitem{sun2014local}
D.~Sun, C.~Liu, and H.~Pfister, ``Local layering for joint motion estimation
  and occlusion detection,'' in \emph{Proc. CVPR}, 2014.

\bibitem{weinzaepfel2013deepflow}
P.~Weinzaepfel, J.~Revaud, Z.~Harchaoui, and C.~Schmid, ``Deepflow: Large
  displacement optical flow with deep matching,'' in \emph{Proc. ICCV}, 2013.

\bibitem{guney2016deep}
F.~G{\"u}ney and A.~Geiger, ``Deep discrete flow,'' in \emph{Proc. ACCV}, 2016.

\bibitem{bai2016exploiting}
M.~Bai, W.~Luo, K.~Kundu, and R.~Urtasun, ``Exploiting semantic information and
  deep matching for optical flow,'' in \emph{Proc. ECCV}, 2016.

\bibitem{sevilla2016optical}
L.~Sevilla-Lara, D.~Sun, V.~Jampani, and M.~J. Black, ``Optical flow with
  semantic segmentation and localized layers,'' in \emph{Proc. CVPR}, 2016.

\bibitem{ranjan2017optical}
A.~Ranjan and M.~J. Black, ``Optical flow estimation using a spatial pyramid
  network,'' in \emph{Proc. CVPR}, 2017.

\bibitem{vvaquero2017flow}
V.~Vaquero, G.~Ros, F.~Moreno-Noguer, A.~M. Lopez, and A.~Sanfeliu, ``Joint
  coarse-and-fine reasoning for deep optical flow,'' in \emph{Proc. ICIP},
  2017.

\bibitem{Menze2015CVPR}
M.~Menze and A.~Geiger, ``Object scene flow for autonomous vehicles,'' in
  \emph{Proc. CVPR}, 2015.

\bibitem{jason2016back}
J.~Y. Jason, A.~W. Harley, and K.~G. Derpanis, ``Back to basics: Unsupervised
  learning of optical flow via brightness constancy and motion smoothness,'' in
  \emph{Proc. ECCV Workshops}, 2016.

\bibitem{Meister2018Unflow}
S.~Meister, J.~Hur, and S.~Roth, ``{UnFlow}: Unsupervised learning of optical
  flow with a bidirectional census loss,'' in \emph{Proc. AAAI}, 2018.

\bibitem{dong2016image}
C.~Dong, C.~C. Loy, K.~He, and X.~Tang, ``Image super-resolution using deep
  convolutional networks,'' \emph{IEEE Transactions on Pattern Analysis and
  Machine Intelligence}, vol.~38, no.~2, pp. 295--307, 2016.

\bibitem{zeiler2010deconvolutional}
M.~D. Zeiler, D.~Krishnan, G.~W. Taylor, and R.~Fergus, ``Deconvolutional
  networks,'' in \emph{Proc. CVPR}, 2010.

\bibitem{vaquero2017deconvolutional}
V.~Vaquero, I.~del Pino, F.~Moreno-Noguer, J.~Sol\`a, A.~Sanfeliu, and
  J.~Andrade-Cetto, ``Deconvolutional networks for point-cloud vehicle
  detection and tracking in driving scenarios,'' in \emph{Proc. ECMR}, 2017.

\bibitem{zheng2015conditional}
S.~Zheng, S.~Jayasumana, B.~Romera-Paredes, V.~Vineet, Z.~Su, D.~Du, C.~Huang,
  and P.~H. Torr, ``Conditional random fields as recurrent neural networks,''
  in \emph{Proc. CVPR}, 2015.

\bibitem{wu2017squeezeseg}
B.~Wu, A.~Wan, X.~Yue, and K.~Keutzer, ``Squeezeseg: Convolutional neural nets
  with recurrent crf for real-time road-object segmentation from 3d lidar point
  cloud,'' \emph{arXiv}, 2017.

\bibitem{vedaldi15matconvnet}
A.~Vedaldi and K.~Lenc, ``Matconvnet -- convolutional neural networks for
  matlab,'' in \emph{Proc. {ACM} Multimedia}, 2015.

\bibitem{he2015delving}
K.~He, X.~Zhang, S.~Ren, and J.~Sun, ``Delving deep into rectifiers: Surpassing
  human-level performance on imagenet classification,'' in \emph{Proc. ICCV},
  2015.

\bibitem{zweig17interponet}
S.~Zweig and L.~Wolf, ``Interponet, a brain inspired neural network for optical
  flow dense interpolation,'' in \emph{Proc. CVPR}, 2017.

\end{thebibliography}

\balance

\end{document}